\documentclass[conference]{IEEEtran}
\IEEEoverridecommandlockouts

\usepackage{cite}
\usepackage{amsmath,amssymb,amsfonts}
\usepackage{graphicx}
\usepackage{subfig}
\usepackage{booktabs}
\usepackage{siunitx}
\usepackage{xcolor}
\usepackage{url}
\usepackage{cleveref}
\crefname{figure}{Fig.}{Figs.}
\Crefname{figure}{Fig.}{Figs.}
\crefname{table}{Table}{Tables}
\Crefname{table}{Table}{Tables}
\crefname{section}{Section}{Sections}
\Crefname{section}{Section}{Sections}
\crefname{equation}{Eq.}{Eqs.}
\Crefname{equation}{Eq.}{Eqs.}

\begin{document}
\bstctlcite{IEEEexample:BSTcontrol}

\title{From Tabletop to Walking Bot: Deploying a VLA Model on a Legged Manipulator Robot}

\author{%
\IEEEauthorblockN{Giovanni Minelli, Antonio Langella, Giovanni B. Dessy, Lorenzo Amatucci,\\
Giulio Turrisi, Victor Barasuol, Claudio Semini}
\IEEEauthorblockA{Dynamic Legged Systems Lab, Istituto Italiano di Tecnologia (IIT)\\
Correspondence: giovanni.minelli@iit.it}
}

\maketitle

\begin{abstract}
Vision-language-action (VLA) models are typically pretrained on broad, diverse datasets and subsequently fine-tuned on a comparatively small dataset tailored to a specific target robot. Such adaptation recipe is now widely established, achieving its most mature results on fixed-base arms due to their static base dynamics and consistent reference frame. Mounting the same arm on a quadruped extends the workspace to the entire floor, but fundamentally changes the embodiment: rather than controlling the arm in isolation, the policy must jointly coordinate base navigation, body posture, and end-effector motion. In this work, we study the practical demands of this embodiment change across data collection, model adaptation, and inference design to achieve closed-loop loco-manipulation on a Unitree Go2 with an AgileX Piper arm.
% Demonstrations are teleoperated through an 11D loco-manipulation command interface, and the fine-tuned policy is then deployed through that same interface and same robot-side controllers, so no retargeting stage is inserted between model and robot.
We report the choices that made this real-robot deployment feasible, namely model variant, action horizon, and demonstration conventions, together with a study tailored to a pick-and-place task, separating data coverage from model capacity. We show that expanding demonstration coverage reduces held-out end-effector error by 30\%, and incorporating secondary skills (such as pushing rather than grasping) introduces new behavioral modes without degrading baseline performance. Finally, closed-loop real-world experiments show how dataset composition shapes success and failure, validating the pipeline for more complex loco-manipulation.
\end{abstract}


\section{Introduction}
\label{sec:introduction}

Adapting a vision-language-action (VLA) model to a new robot follows a \textit{de-facto} standard procedure: a policy pretrained on large multi-robot manipulation datasets is fine-tuned on a modest number of demonstrations of the target task, each paired with a language instruction, and the resulting policy is queried at runtime for short sequences of actions. The procedure has proven effective on fixed-base manipulators and, more recently, on wheeled mobile platforms \cite{fu2024mobile,MOLMO}.

\begin{figure}[t]
\vspace{-0.7em}
\centering
\subfloat[Approach]{\includegraphics[width=.33\linewidth]{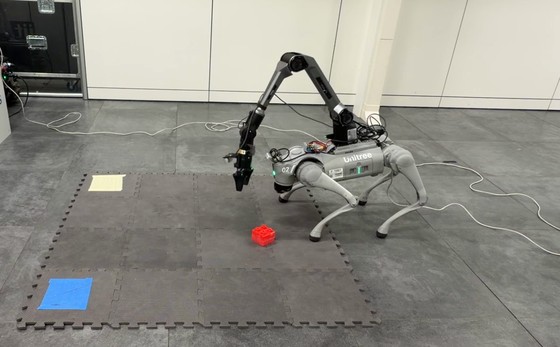}}\hfill
\subfloat[Grasp]{\includegraphics[width=.33\linewidth]{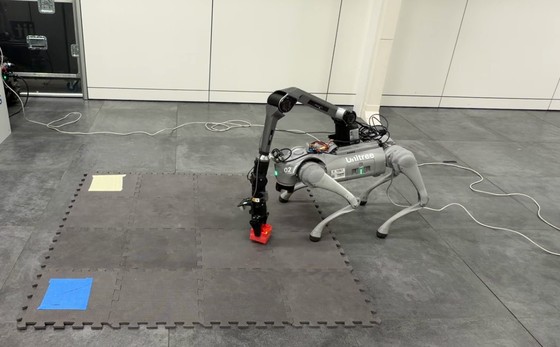}}\hfill
\subfloat[Transport]{\includegraphics[width=.33\linewidth]{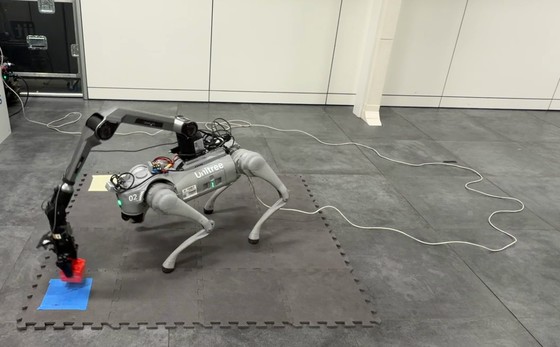}}
\caption{Frames from a rollout of the deployed policy on the Go2 quadruped carrying the Piper arm. Body pitch, base motion, and arm motion are commanded through the same interface used during demonstration.}
\label{fig:rollout}
\vspace{-1em}
\end{figure}

Extending from fixed-base to legged manipulators changes two things at once. The first is the workspace of the robot. On a fixed-base setup, everything the robot's policy can reach is set by the arm, and an object placed beyond it is simply unreachable. Mounted on a quadruped, the same arm can reach anywhere the robot can walk, so the policy must first move the body to bring the target within the arm's workspace. The second thing is the commanded action space of the policy. A policy on a fixed-base arm emits arm commands alone, whereas a policy on a legged manipulator must additionally command base motion and body posture while the base keeps its balance. Reaching for an object is therefore an inherently coordinated behaviour in which stepping, body pitch, and arm motion are decided together.

Both changes matter at deployment. In our instantiation, instead of joint space commands, the policy output is expressed in the quantities that a pretrained locomotion controller expects, so collection and deployment share the same physical factors, such as control delays, contact, and balance dynamics. Collecting the required coordinated demonstrations is itself not straightforward: existing legged teleoperation systems either instrument the operator with wearable motion capture or delegate part of the base motion to an autonomous policy 
\cite{TeleMAN,honerkamp2024zero}. In such cases, decisions about base placement are absent from the recorded data.

This extended abstract reports initial work toward carrying a pretrained VLA across this change of embodiment and what the resulting performance appears limited by. The next sections describe: a collection setup in which one operator commands base and arm concurrently through an interface later shared with the policy; the training and inference choices that made the model usable on hardware; and a controlled variation of the training data that separates additional examples of a known behaviour from examples of a new one. We close with an initial deployment on the physical robot and outline how this approach can be expanded toward broader loco-manipulation abilities.

\section{Data Collection}
\label{sec:collection}

The platform adopted consists of a Unitree Go2 quadruped carrying an AgileX Piper L arm with a parallel gripper, a wrist-mounted RealSense D435 camera, and a D405 mounted on the quadruped's head that provides the forward view of the scene. The target loco-manipulation task consists in picking up a plastic cube and placing it on a colored target area. Instructions are drawn from a template set that varies the verb, the instruction phrasing, and the object color, so the policy is not fitted to a single sentence. A further subset of demonstrations reaches the same goal by pushing the cube rather than grasping it, labelled with goal-oriented phrasings that contain no \textit{grasping} verb.

A single operator collects each demonstration. A second Piper arm serves as a haptic device in a leader-follower pair, whose kinematic configuration is mirrored by the follower arm. Two handheld controllers supply the remaining degrees of freedom: body pitch, yaw, gripper toggling, episode markers, and base velocity. One controller is mounted on the leader-arm handle. The same hand therefore can drive arm motion, gripper, and posture, while the other hand commands base velocity. A bilateral teleoperation coupling between leader and follower arm renders contact cues without force sensors.  Driving base and arm concurrently yields coordinated reaching and body placement, avoiding less natural motions derived from interleaving a discrete navigation phase with a subsequent manipulation phase. Demonstrations are recorded at \SI{5}{Hz}, aligned to the slower \SI{10}{Hz} head camera against the \SI{22}{-}\SI{27}{Hz} wrist camera.

\begin{figure}[t]
\centering
\includegraphics[width=0.95\linewidth]{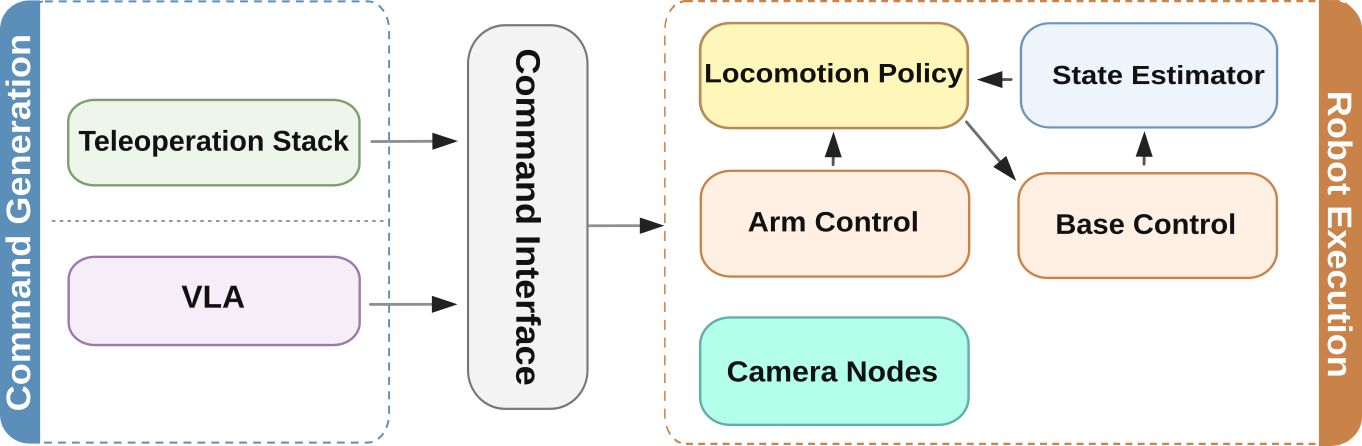}
\caption{Collection and deployment share the robot-side perception, locomotion, and arm-control modules. At deployment, VLA inference replaces the leader-side interface.}
\label{fig:architecture}
\vspace{-1em}
\end{figure}

% In our framework,
The operator and the policy issue the same command,
\begin{equation}
\mathbf{a} = [v_x,v_y,\omega_z,\theta,q_{\mathrm{arm}}^{1\dots6},g]^{\top}\in\mathbb{R}^{11},
\label{eq:action}
\end{equation}
comprising, in order, body-frame base twist, body-pitch reference, six absolute arm joint position targets, and the gripper command. The robot-side stack behind this 11D command interface is identical during collection and deployment, with base state supplied by the onboard MUSE estimator \cite{MUSE}, base commands entering an arm-aware reinforcement-learning locomotion policy derived from BinWalker \cite{turrisi26littercollection}, and arm commands executed by the joint-trajectory controller (see \cref{fig:architecture}). Representing actions as body-level commands, rather than leg joint references, keeps what the policy must learn at the level of coordinated placement decisions. Moreover, using the same controller for operator and high level policy commands keeps collection and execution consistent. A policy trained on these labels can therefore take the operator's place at this command boundary, with no inverse kinematics or controller retargeting between model and robot.

The action labels follow four conventions that make the learned behaviour directly deployable. The base action stores the twist the operator commanded rather than the velocity the robot achieved, so the learned quantity coincides with the quantity the controller expects. Body pitch is recorded in the reference space the controller expects, which allows the policy output to be forwarded without rescaling. Arm targets are absolute joint-position targets, so each prediction overwrites the previous one and errors do not accumulate as they would with relative delta commands integrated over time. The gripper actions encode open-or-closed intent rather than measured gripper opening, which would otherwise record the grasped cube's width instead of the decision to close.

\section{Training and Inference}
\label{sec:training}

We build on the MolmoAct2 family \cite{MOLMO}, chosen for its open weights and transparency on training recipe. The models adopt a flow-matching expert action head, producing a chunk of actions per observation. The family also offers a reasoning variant with scene-depth prediction. While it enables extended reasoning, the resulting inference latency is too high for closed-loop control of the quadruped. We therefore fine-tune and deploy the plain model with depth reasoning disabled, using rank-512 low-rank adaptation of the vision-language backbone together with a fully trained action expert on a single GPU RTX 6000 Pro (\SI{96}{GB} VRAM).

The action horizon is set empirically to balance outdated plans, inference delays, and data characteristics. A common choice is to predict one second of action, discretized into five \SI{200}{ms} steps at \SI{5}{Hz}, matching data collection frequency. In our system this horizon produced jerky and imprecise motions because each plan was replaced before the motion could settle. Conversely, fifteen-step plans continued to execute commands that no longer matched the current camera view. Empirically, ten actions at \SI{5}{Hz} proved a workable middle ground. The arm receives the whole action chunk, a sequence of predicted actions, as one multi-point trajectory to avoid goal preemption, while the base, which accepts only instantaneous commands, is served by a \SI{50}{Hz} keep-alive thread holding the absolute pitch reference and zeroing velocity between inferences.

\section{Experiments}
\label{sec:experiments}

Building on the deployment setup established above, we examine how far the instructed pick-and-place behaviour is limited by data. We address it by training on three nested datasets while keeping every other setting fixed. We denote them as: \textbf{D1}, contains 131 grasp episodes; \textbf{D2}, adds grasp episodes from a wider range of starting positions, approach directions, and stances, for 233 episodes; \textbf{D3}, further adds 25 push episodes for a total of 258. The held-out set is reported separately per task: 26 grasp episodes, in-distribution for all three models therefore providing fair comparison, and 5 push episodes, in-distribution for D3 alone. Because mean joint error saturates and ranks checkpoints poorly, we report end-effector error obtained through the kinematic chain, expressed in millimeters and therefore directly comparable to the positioning accuracy required to grasp and place the cube, alongside gripper open/close accuracy and pitch error in degrees.

\begin{table}[t]
\centering
\footnotesize
\caption{Held-out action-prediction error. Grasp is in-distribution for all models and is the common comparison; push is in-distribution only for D3, so the D1 and D2 push rows are an out-of-distribution reference.}
\label{tab:results}
\begin{tabular}{@{}lcccccc@{}}
\toprule
 & \multicolumn{3}{c}{Grasp (26 ep.)} & \multicolumn{3}{c}{Push (5 ep.)} \\
\cmidrule(lr){2-4}\cmidrule(lr){5-7}
Model & EE Tip & Pitch & Gripper & EE Tip & Pitch & Gripper \\
(episodes) & (mm) & (\si{\degree}) & (\% acc.) & (mm) & (\si{\degree}) & (\% acc.) \\
\midrule
D1 (131) & 68.7 & 5.01 & 94 & 77.5 & 8.53 & 50 \\
D2 (233) & 48.2 & 2.24 & \textbf{99} & 74.3 & 2.36 & 27 \\
D3 (258) & \textbf{47.8} & \textbf{1.87} & 98 & \textbf{67.0} & \textbf{1.63} & \textbf{57} \\
\bottomrule
\end{tabular}
\end{table}

\begin{figure}[t]
\centering
\includegraphics[width=\linewidth]{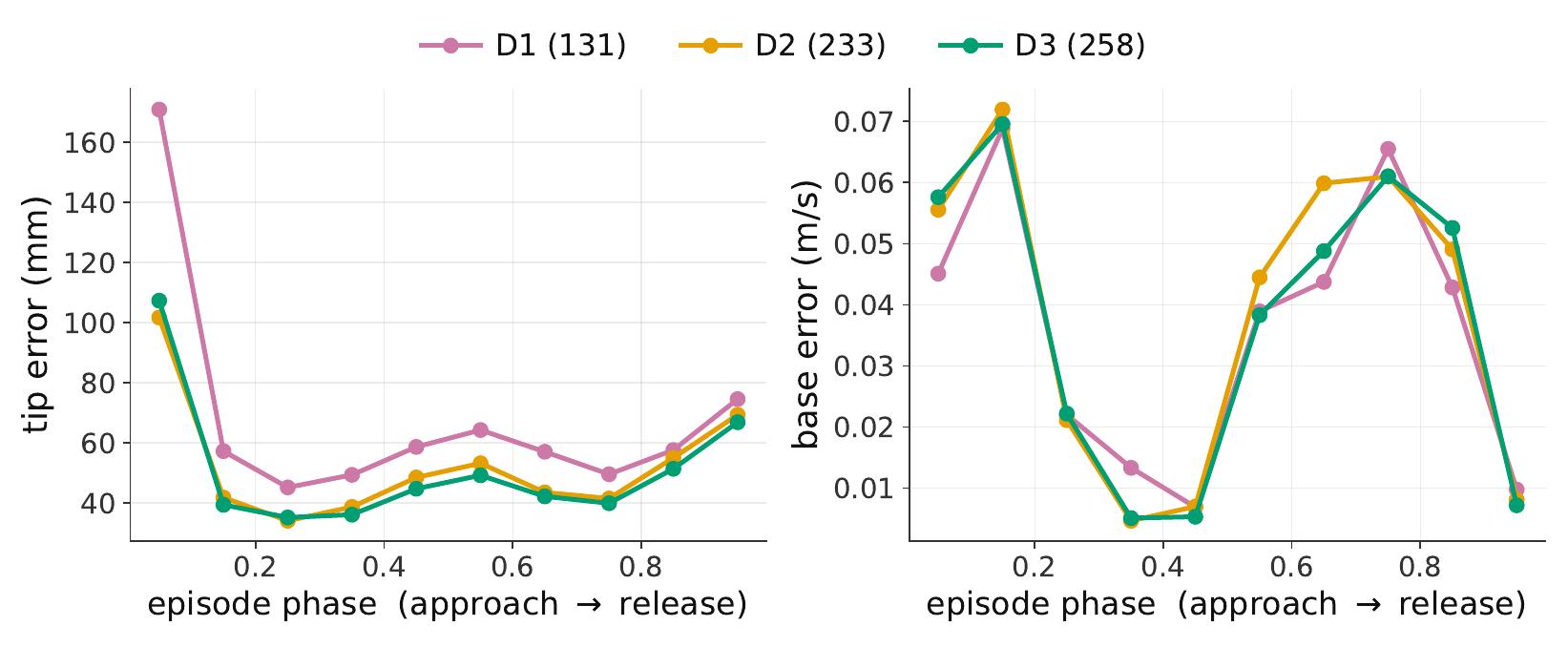}
\caption{End-effector and base error across the episode for the three training sets. The largest reduction is at the initial \textit{approach} phase.}
\label{fig:phase}
\vspace{-1em}
\end{figure}

Table \ref{tab:results} summarizes held-out action-prediction error for grasp and push. On the common grasp comparison, end-effector error falls from \SI{68.7}{mm} to \SI{48.2}{mm} between D1 and D2 and remains essentially unchanged at \SI{47.8}{mm} for D3. Pitch error falls from \SI{5.01}{\degree} to \SI{1.87}{\degree}, indicating that the additional demonstrations improve the body posture required to bring the cube into the arm workspace. Gripper accuracy rises from 94\% to near saturation. On the push set, D3 attains \SI{67.0}{mm} and \SI{1.63}{\degree} with 57\% gripper accuracy, against \SI{74.3}{mm}, \SI{2.36}{\degree}, and 27\% for D2. \cref{fig:phase} decomposes grasp error along the episode, from initial approach through cube placement to release.

Further analysis reveals that the D1-to-D2 gain stems from broader coverage of an existing behaviour. \cref{fig:phase} localises this improvement to the initial approach, where the policy must commit to a body placement before the cube is well-framed. A densely sampled analysis attributes roughly \SI{65}{mm} of the error reduction to this phase alone, alongside an approximate 65\% decrease in mean pitch error. The D2-to-D3 comparison differs in nature, as the models vary by the introduction of a new behaviour rather than the volume of an existing one. Since pushing is absent from D1 and D2, their corresponding push columns reflect out-of-distribution generalisation rather than scaling. D3 successfully acquires this pushing behaviour at no measurable cost to grasping, with grasping error remaining unchanged between D2 and D3. % In short, broader coverage of grasping improves grasping, while only demonstrations of pushing teach pushing.

We also verified that the improvement is attributable to data rather than adaptation. Increasing adaptation capacity from low-rank to full fine-tuning did not move validation error away from the plateau observed on the D1 dataset, whereas enlarging the dataset did. At the demonstration counts available here, data coverage limits performance more than model capacity does.

\section{Hardware Deployment}
\label{sec:deployment}

Offline results compare predictions against recorded demonstrations, yet on their own they are insufficient to judge hardware deployment readiness: they cannot show whether errors compound once the policy closes the loop through the real robot. We therefore ran the trained policies on the physical Go2 and Piper. % without modifying the interface used during collection, keeping on-robot trials as part of best performing checkpoint selection.
Given the two camera streams, robot state, and an instruction, the policy approaches the cube, pitches the body to bring the object within the arm workspace, closes the gripper, transports the cube, and releases it at the target. \cref{fig:rollout} shows frames from such an execution. The preliminary result is qualitative rather than a success-rate study, but it closes the path from teleoperated demonstrations through fine-tuning to autonomous loco-manipulation on hardware.

The failures observed on the robot are informative because they match the offline analysis. When the cube leaves the camera view in which the robot can act, the policy does not reliably reorient the body to observe it, and when the commanded base motion is too conservative the arm begins its movement with the target base at the edge of the workspace. Both failures concern initial body placement, the initial-approach phase that \cref{fig:phase} shows as least accurate and most improved by additional data. Dataset composition therefore shapes success and failure directly: navigation cues and spatial coverage decide where the rollout succeeds and where it breaks.

\section{Conclusion}
We presented practical steps to transfer a pretrained VLA across a change of embodiment, from a fixed-base arm to a legged manipulator, via a shared demonstration and deployment interface validated in closed-loop real-robot execution. Our task-tailored study shows that broader grasp demonstrations reduce held-out end-effector error by 30\%, primarily by refining the base placement and posture commands unique to the new platform. Furthermore, \textit{push} demonstrations introduce a distinct interaction mode without degrading existing grasping performance, confirming that targeted data can broaden behavioral capabilities with no degradation to the base grasping skill. Ultimately, this validated approach provides a reliable foundation for developing more complex loco-manipulation skills.

\bibliographystyle{IEEEtran}
\bibliography{ref}

\end{document}